\documentclass[10pt,twocolumn]{article}
\usepackage[margin=2cm]{geometry}
\usepackage{graphicx}
\usepackage{amsmath,amssymb}
\usepackage{booktabs}
\usepackage{hyperref}
\usepackage{natbib}
\usepackage{xcolor}
\usepackage{caption}

\title{Cross-Species RSA Reveals Conserved Early Visual Alignment\\but Divergent Higher-Area Rankings Across\\Human fMRI and Macaque Electrophysiology}

\author{Nils Leutenegger\\
Independent Researcher, Switzerland\\
\texttt{github.com/nilsleut}}

\date{}

\begin{document}
\maketitle

\begin{abstract}
\noindent\textbf{Correction (August 2026).} An evaluation-mode defect in the shared
feature-extraction pipeline affected the predictive-coding and STDP conditions. It
applies to both sides of every cross-species comparison here, since the human values
are reprinted from the companion study and the macaque values use the same
checkpoints. A single-seed re-evaluation with repaired checkpoints leaves finding~(2)
supported for STDP but not for predictive coding, and changes the V4 and IT orderings.
The other conditions and the ResNet-50 capacity control change by at most $0.0034$.
See the correction note following this abstract.

\medskip

Does the relationship between learning rules and brain alignment generalize across species?
We extend our prior finding that untrained CNNs match backpropagation at human V1 by testing the same five learning rules against macaque electrophysiology data.
The rules are backpropagation (BP), feedback alignment (FA), predictive coding (PC), spike-timing-dependent plasticity (STDP), and an untrained random-weights baseline.
The macaque data come from two publicly available datasets: MajajHong2015 (V4/IT, 3{,}200 stimulus presentations, 88/168 neurons) and FreemanZiemba2013 (V1/V2, 135 stimuli, 102/103 neurons).
Using Representational Similarity Analysis (RSA) with identical model weights from our human study, we find that:
(1)~all models achieve substantially higher alignment with macaque early visual cortex ($\rho = 0.15$--$0.30$ at V1/V2) than with human fMRI ($\rho = 0.01$--$0.08$), consistent with the higher signal-to-noise ratio of electrophysiology;
(2)~STDP and PC produce the highest macaque V1/V2 alignment ($\rho \approx 0.30$ and $0.28$), consistent with their leading position among trained rules in human V1;
(3)~at IT, learning rule rankings show no detectable correlation across species (Kendall's $\tau = 0.00$, $p = 1.00$), though this null result is expected given that $n = 5$ provides statistical power only at $\tau = \pm 1.0$, and is further confounded by stimulus set differences;
(4)~a pretrained ResNet-50 baseline (ImageNet) achieves $\rho = 0.25$ at macaque IT, substantially above all custom CNN conditions ($\rho = 0.07$--$0.14$), suggesting that IT alignment in our setup is limited by model capacity and training data rather than by the learning rule.
Noise ceilings, multi-seed variability (5 seeds), and a stimulus-control analysis are reported.
These results demonstrate that early visual alignment is robust across species and measurement modalities, while higher-area alignment is modulated by model capacity and stimulus domain.
\end{abstract}

\section*{Correction note (August 2026)}

\noindent The human V1 results of this paper, the untrained network above backpropagation and the human V1 ranking of learning rules, are not robust to the analysis choices examined in ``Analysis choices decide whether untrained networks beat trained ones at V1: a multiverse analysis of model--brain RSA'' arXiv:2611.15533. In the model set re-tested there, the untrained-minus-backpropagation difference at human V1 is $-$0.011 [$-$0.017, $-$0.005] with standardised model units. The macaque comparisons and the cross-species conclusions were not re-tested. The text below is unchanged.

\noindent\textbf{The defect propagates from the companion study.} The
predictive-coding and STDP model classes in the shared implementation override
\texttt{eval()} with a no-op. Feature extraction calls \texttt{eval()} before
computing activations, so for those two conditions the batch-normalisation layers
stayed in training mode and normalised each evaluation batch by its own statistics,
while the random, backpropagation and feedback-alignment conditions used their stored
running statistics. This is a rule-correlated difference in the evaluation pipeline,
not in the learning rules.

\begin{figure*}[t]
\centering
\includegraphics[width=\textwidth]{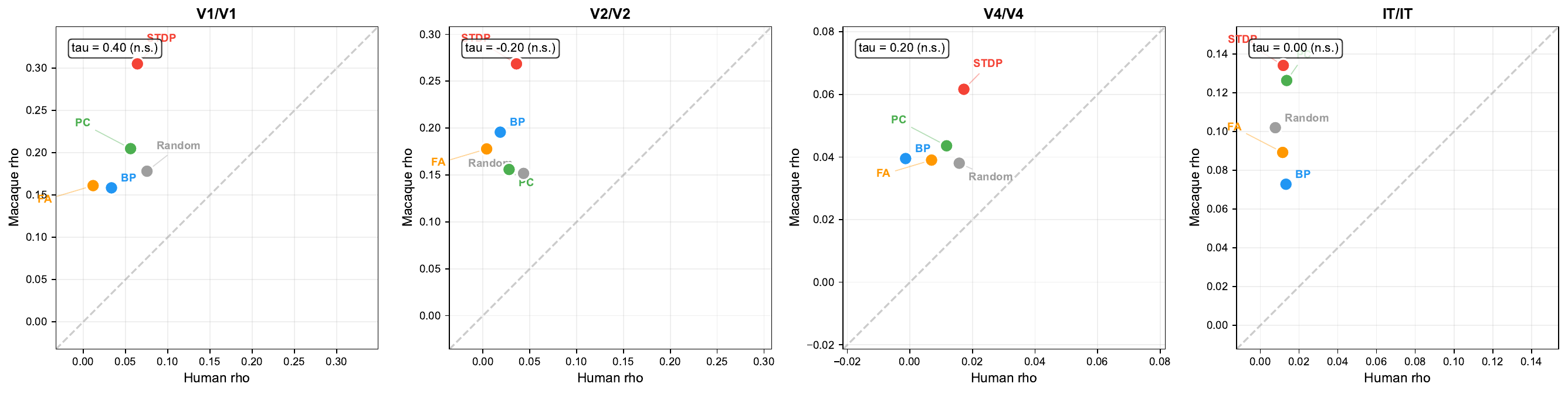}
\caption{Cross-species ranking comparison.
Each panel shows human $\rho$ (x-axis) vs.\ macaque $\rho$ (y-axis) per learning rule at one brain region.
Kendall's $\tau$ with exact $p$-values from exhaustive permutation over all $5! = 120$ orderings.
At $n = 5$, only $\tau = \pm 1.0$ can reach significance at $\alpha = 0.05$.
All macaque points lie above the identity line, reflecting the electrophysiology SNR advantage.}
\label{fig:scatter}
\end{figure*}

\begin{figure}[t]
\centering
\includegraphics[width=\columnwidth]{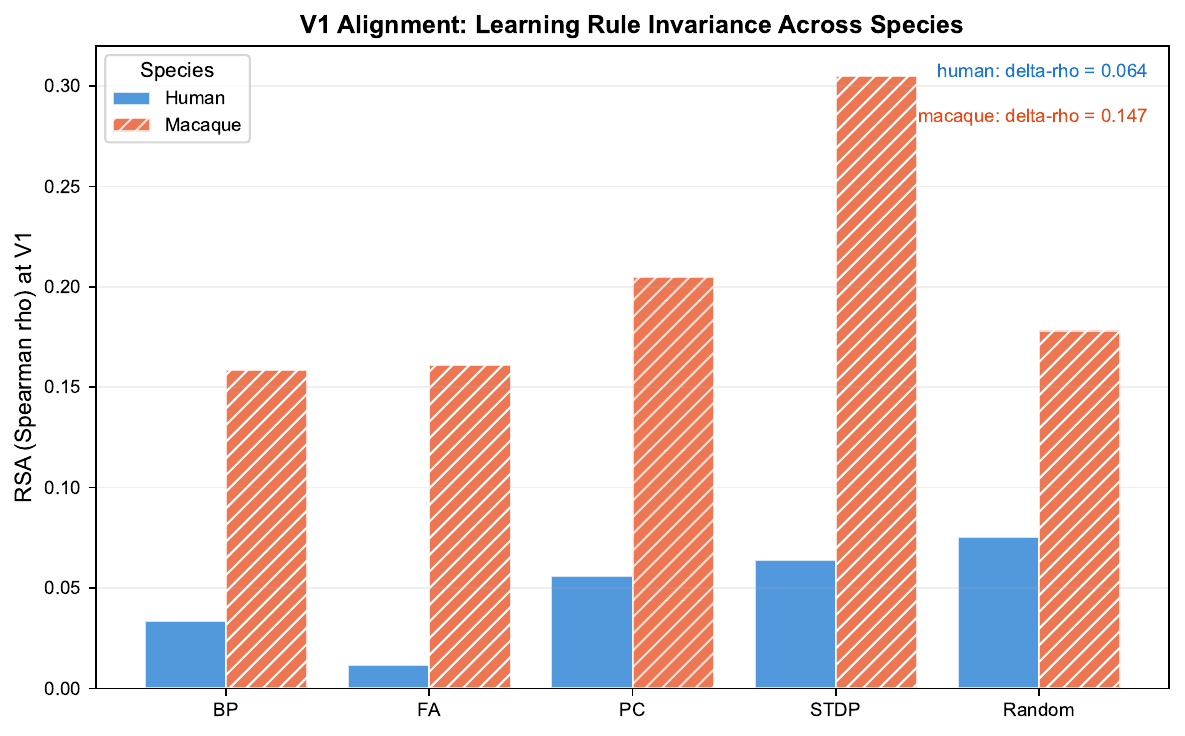}
\caption{V1 alignment per learning rule, grouped by species.
Human $\Delta\rho = 0.064$; macaque $\Delta\rho = 0.147$.
Electrophysiology resolves larger differences between learning rules at V1 than fMRI.}
\label{fig:v1}
\end{figure}

\begin{figure*}[t]
\centering
\includegraphics[width=\textwidth]{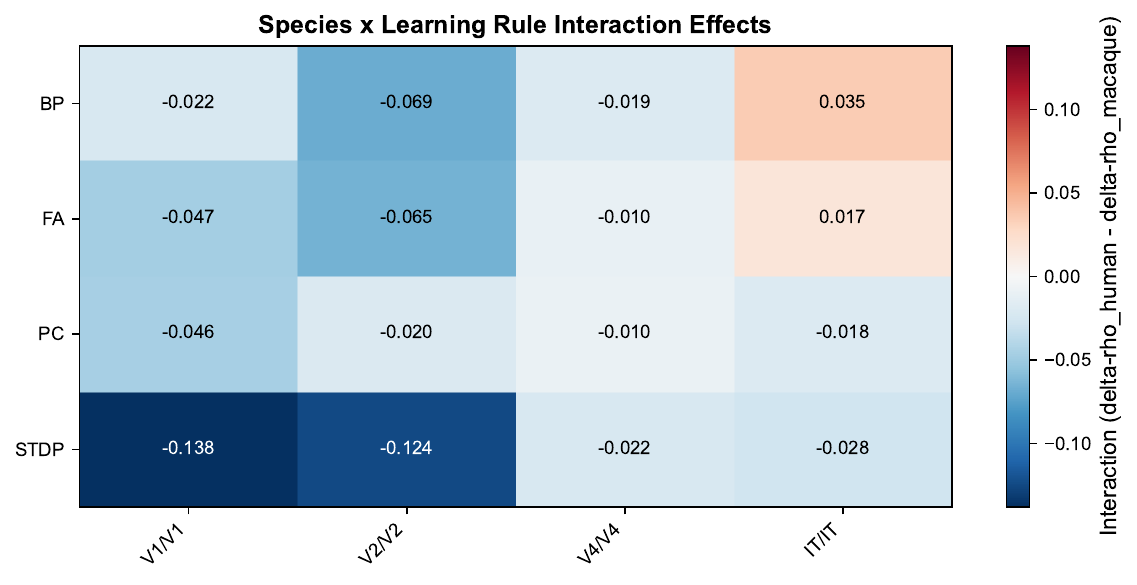}
\caption{Species $\times$ learning rule interaction effects.
Values show $(\Delta\rho_\text{human} - \Delta\rho_\text{macaque})$ where $\Delta\rho = \rho_\text{rule} - \rho_\text{random}$.
Negative values (blue): macaque benefits more from the learning rule than human.
STDP and PC show strong negative interactions at V1/V2.}
\label{fig:interaction}
\end{figure*}

\begin{figure*}[t]
\centering
\includegraphics[width=\textwidth]{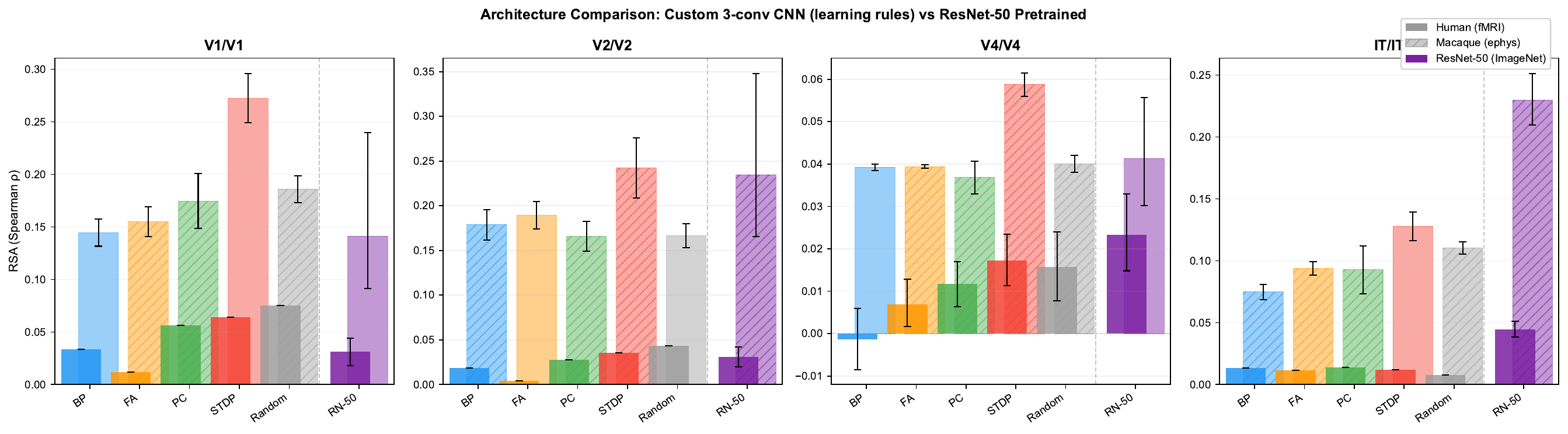}
\caption{Architecture comparison: custom 3-conv CNN (5 learning rules) vs.\ pretrained ResNet-50.
At V1/V2, ResNet-50 performs comparably to BP/FA but below STDP/PC.
At IT, ResNet-50 ($\rho \approx 0.25$) dramatically outperforms all CNN conditions ($\rho = 0.07$--$0.14$), suggesting that IT alignment scales with model capacity and training data.
Macaque error bars: mean $\pm$ std across 5 seeds.
$^\dagger$STDP IT uses $n = 4$ seeds (seed-0 lacked FC1 weights; see Section~3.1).}
\label{fig:resnet}
\end{figure*}

\begin{figure*}[t]
\centering
\includegraphics[width=\textwidth]{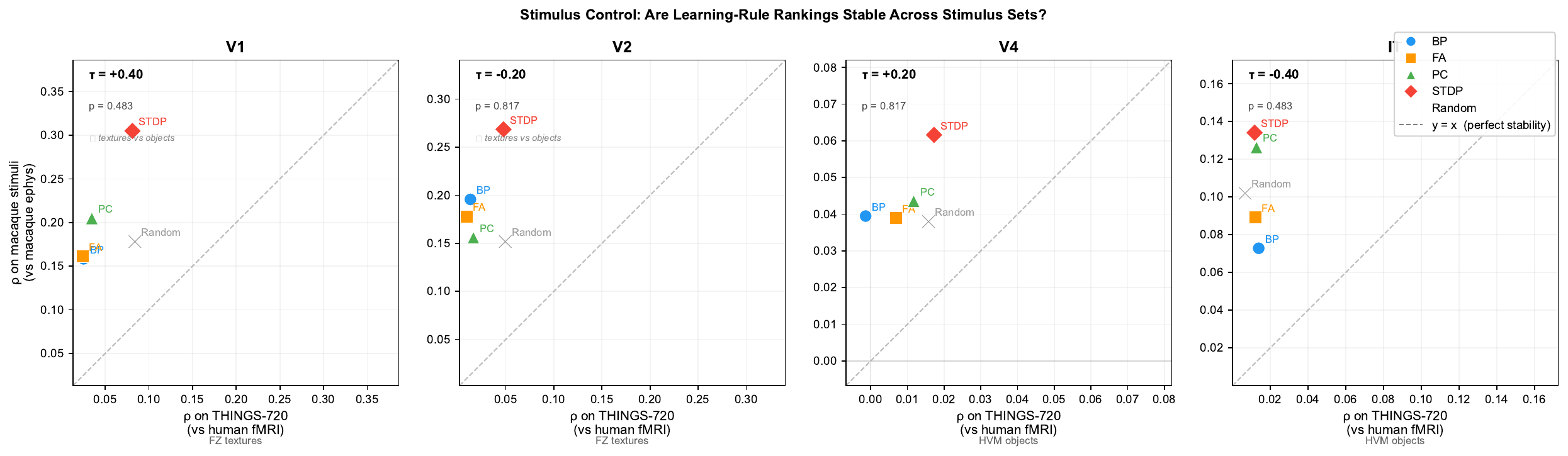}
\caption{Stimulus control: are learning-rule rankings stable across stimulus sets?
Each panel shows model $\rho$ on THINGS-720 (human stimuli, x-axis) vs.\ model $\rho$ on macaque stimuli (y-axis).
Rankings are moderately stable at V1 ($\tau = 0.40$) but weakly inverted at IT ($\tau = -0.40$), suggesting that cross-species IT differences may be partially driven by stimulus domain.}
\label{fig:stimulus}
\end{figure*}

\noindent\textbf{It affects both sides of every comparison in this paper.} The macaque
values here were computed with the affected checkpoints, and the human values are
reprinted from the companion study \citep{leutenegger2025}, which used the same
pipeline. Every cross-species quantity in this paper therefore combines two affected
numbers wherever predictive coding or STDP is involved: the macaque and human profiles
(Fig.~\ref{fig:profiles}), the ranking scatterplots (Fig.~\ref{fig:scatter}), the V1
invariance comparison (Fig.~\ref{fig:v1}), the interaction effects
(Fig.~\ref{fig:interaction}), the architecture comparison (Fig.~\ref{fig:resnet}) and
the stimulus control (Fig.~\ref{fig:stimulus}), together with Sections~4.1--4.6, the
corresponding discussion, contribution~(2) of the introduction, finding~(2) of the
abstract, and the conclusion.

\noindent\textbf{What has been re-computed, and what has not.} We have re-run the
macaque evaluation with repaired checkpoints in the single-seed resolution-control
setting. That establishes the direction and magnitude of the shift but is not a re-run
of this paper's five-seed analysis. Kendall's $\tau$ across species, the noise
ceilings, the stimulus-control analysis and the seed-level variability have \emph{not}
been re-computed, and the numbers printed in the body remain the affected ones. In
that single-seed re-evaluation at $224$\,px, Spearman $\rho$ moves as follows (old
$\to$ repaired):

\vspace{0.4em}
\begingroup\scriptsize\setlength{\tabcolsep}{3.2pt}
\begin{tabular}{@{}lcccc@{}}
\toprule
 & V1 & V2 & V4 & IT \\
\midrule
STDP     & $.305\!\to\!.266$ & $.268\!\to\!.211$ & $.062\!\to\!.055$ & $.134\!\to\!.149$ \\
PC       & $.210\!\to\!.178$ & $.155\!\to\!.165$ & $.045\!\to\!.040$ & $.125\!\to\!.096$ \\
Random   & $.178$            & $.152$            & $.038$            & $.102$ \\
FA       & $.161\!\to\!.158$ & $.178\!\to\!.177$ & $.039$            & $.089$ \\
BP       & $.158\!\to\!.156$ & $.196$            & $.040\!\to\!.037$ & $.073\!\to\!.072$ \\
\midrule
ordering & changed & unchanged & changed & changed \\
\bottomrule
\end{tabular}
\endgroup
\vspace{0.6em}

\noindent\textbf{Which claims survive and which do not.} Finding~(2), that STDP and
predictive coding produce the highest macaque V1/V2 alignment, is \emph{half}
supported. STDP retains a clear lead at both V1 and V2 after repair. Predictive coding
does not: at V1 it falls to $\rho = 0.178$, level with the untrained baseline, so it
can no longer be described as leading there. At V2 the ordering is unchanged, but
predictive coding already ranked below backpropagation and feedback alignment in that
region before the repair. At V4 and IT the orderings change and none of the rankings
reported for those regions should be used. Finding~(1), that macaque alignment
substantially exceeds human fMRI alignment, is unaffected. So are finding~(3), the
statistical-power argument that Kendall's $\tau$ at $n=5$ can only reach significance
at $\tau = \pm 1.0$; finding~(4), the ResNet-50 capacity control at macaque IT; and the
stimulus-domain confound between FreemanZiemba textures and HVM objects. The random,
backpropagation and feedback-alignment conditions and the ResNet-50 control change by
at most $0.0034$, which alters no ordering among them.

\noindent\textbf{On checkpoints.} Training in this implementation is deterministic, so
re-running it reproduces the same weights. The regenerated seed-0 STDP checkpoint is
therefore bit-identical to the original file in its convolutional weights; what changed
is that it now serialises the full state dictionary rather than the three convolutional
tensors alone. Section~3.1 notes that the original file lacked its fully connected
readout, so any evaluation reloading it used a randomly initialised FC1; that caveat
stands, and the regenerated file removes it. The original file also lacked its
batch-normalisation running statistics. Under the defective pipeline this had no
effect, because batch normalisation was operating on batch statistics and never
consulted the stored ones; it matters only for evaluation with the repaired
implementation, where those statistics are used.

\vspace{0.4em}
\noindent A full re-computation of this paper's analyses is in preparation. A follow-up
study \citep{leuteneggerERC} additionally shows that alignment in this setting depends
strongly on the resolution at which models are evaluated, which was held fixed at
$224$\,px throughout this paper. The repaired implementation and all re-run results are
at \url{https://github.com/nilsleut/evaluation-resolution-rsa}.

\bigskip

\textbf{Keywords:} cross-species, representational similarity analysis, visual cortex, learning rules, macaque electrophysiology, fMRI, Brain-Score

\section{Introduction}

A central finding of computational neuroscience is that deep neural networks (DNNs) trained with backpropagation reproduce the representational hierarchy of the primate ventral visual stream, with early layers matching V1 and deeper layers matching IT cortex \citep{yamins2016,schrimpf2020}.
However, most model--brain comparisons have been conducted within a single species (typically human fMRI), leaving open the question of whether model--brain alignment patterns generalize across species and measurement modalities.

In a companion study \citep{leutenegger2025}, we showed that an untrained random-weights CNN exceeds backpropagation at human V1 ($\rho = 0.076$ vs.\ $0.034$, $p < 0.001$), that STDP and PC preserve more V1-like structure than BP, and that all rules converge at IT.
These findings raise a natural cross-species question: does the same pattern hold when evaluated against macaque electrophysiology, which offers single-neuron resolution and substantially higher signal-to-noise ratios than fMRI?

We address this question by evaluating the same five learning rule conditions, using the identical trained weights from our human study, against two macaque neural datasets from the Brain-Score ecosystem \citep{schrimpf2020}: MajajHong2015 \citep{majaj2015} for V4 and IT (multi-electrode array recordings, 3{,}200 presentations), and FreemanZiemba2013 \citep{freeman2013} for V1 and V2 (single-unit recordings, 135 texture stimuli).
As a capacity control, we additionally evaluate a pretrained ResNet-50 (ImageNet) on all macaque datasets.

Our contributions are:
(1)~the first systematic cross-species RSA comparison of four biologically plausible learning rules plus an untrained baseline, using identical model weights across species;
(2)~evidence that early visual alignment (V1/V2) is qualitatively conserved across human fMRI and macaque electrophysiology, with local learning rules (STDP, PC) outperforming BP in both species and achieving the highest macaque V1 alignment overall;
(3)~evidence that higher-area (IT) rankings show no detectable cross-species correspondence in our small-architecture setting, with a stimulus-control analysis showing that ranking differences are partially attributable to stimulus domain;
(4)~a ResNet-50 capacity control showing that IT alignment scales with model capacity and training data richness, suggesting that the IT convergence observed in our companion study may reflect architectural and data limitations rather than a fundamental property of learning rules.

\section{Related Work}

\textbf{Cross-species model--brain comparisons.}
The Brain-Score framework \citep{schrimpf2020} provides standardized benchmarks for both human and macaque neural data, enabling systematic cross-species evaluation.
\citet{yamins2016} showed that task-optimized DNNs predict macaque IT responses, but did not compare multiple learning rules.
\citet{majaj2015} demonstrated that linear readouts from IT firing rates predict human object recognition, establishing a functional link between macaque IT and human behavior.
To our knowledge, no study has systematically compared the cross-species alignment of multiple learning rules on the same architecture.

\textbf{Learning rules and brain alignment.}
Our companion study \citep{leutenegger2025} provides a detailed comparison of BP, FA, PC, STDP, and an untrained baseline against human fMRI; we refer readers there for a full review of biologically plausible learning rules and their relationship to brain data.

\section{Methods}

\subsection{Models}

All five conditions use the identical architecture and weights from \citet{leutenegger2025}: a custom 3-layer CNN (Conv1: 32 filters, Conv2: 64, Conv3: 128, FC1: 512, FC2: 10) trained on CIFAR-10 for 40 epochs.
The five learning rules are: Backpropagation (BP), Feedback Alignment (FA; \citealt{lillicrap2016}), Predictive Coding (PC; \citealt{whittington2017}), STDP \citep{bi1998}, and Random Weights (untrained).
Model weights are loaded from the companion study's checkpoints; results are averaged across 5 random seeds trained with identical hyperparameters.
One exception: the seed-0 STDP checkpoint from the companion study saved only convolutional weights (FC1 was randomly initialized), so STDP results at IT (FC1-based) use the mean across seeds 1--4 only; V1/V2 results (Conv1-based) are unaffected.
For RSA evaluation, all stimuli are resized to $224 \times 224$.

As a capacity control, we evaluate a pretrained ResNet-50 (ImageNet-1K-V2 weights; \citealt{he2016}) with the layer mapping: \texttt{layer1}$\to$V1/V2, \texttt{layer2}$\to$V4, \texttt{layer4}$\to$IT.

\subsection{Neural Datasets}

\textbf{Human fMRI.}
We use the THINGS-fMRI dataset (720 stimuli, 3 subjects) from \citet{leutenegger2025}.
RSA values ($\rho$) for V1, V2, LOC, and IT are taken directly from that study.
To enable cross-species comparison at V4, we additionally computed Conv2$\to$V4 alignment on the same THINGS-fMRI data using the identical RSA pipeline; this ROI was not reported in the companion study.

\textbf{Macaque V4/IT: MajajHong2015.}
Multi-electrode array recordings from macaque V4 (88 neurons) and IT (168 neurons) in response to 3{,}200 presentations of HVM object images \citep{majaj2015}.
Loaded via the Brain-Score API (\texttt{MajajHong2015.public}).
Responses were averaged across repetitions per stimulus, and neural RDMs were computed using correlation distance.

\textbf{Macaque V1/V2: FreemanZiemba2013.}
Single-unit recordings from macaque V1 (102 neurons) and V2 (103 neurons) in response to 135 texture stimuli \citep{freeman2013}.
Loaded via Brain-Score (\texttt{FreemanZiemba2013.public}).
We note that these are texture stimuli, not object images; this stimulus mismatch is addressed in our stimulus-control analysis (Section~\ref{sec:stimulus_control}).

\textbf{Layer-to-region mapping.}
For the custom CNN: Conv1$\to$V1/V2 (FreemanZiemba), Conv2$\to$V4 (MajajHong and THINGS-fMRI), FC1$\to$IT (MajajHong).
For the human data, the mapping is Conv1$\to$V1, Conv1$\to$V2, Conv2$\to$V4, Conv3$\to$LOC, FC1$\to$IT, consistent with the standard ventral-stream correspondence \citep{yamins2016}.

\subsection{RSA Pipeline}

Feature extraction, RDM construction (correlation distance), and RSA scoring (Spearman $\rho$ between upper triangles) follow the identical pipeline from \citet{leutenegger2025}.
Bootstrap 95\% confidence intervals are estimated from $N = 10{,}000$ stimulus resamples.

\textbf{Noise ceilings} are estimated via split-half reliability (splitting neurons into random halves, computing RDM from each, correlating; $N = 100$ splits), corrected with the Spearman--Brown formula.

\subsection{Cross-Species Analysis}

\textbf{Ranking comparison.}
For each brain region, we compute Kendall's $\tau$ between the human and macaque $\rho$-vectors (one $\rho$ per learning rule).
With $n = 5$ learning rules, an exhaustive permutation test over all $5! = 120$ permutations provides exact $p$-values.
We note that at $n = 5$, only $\tau = \pm 1.0$ can reach significance at $\alpha = 0.05$ ($p = 0.0083$); all other $\tau$ values are descriptive.

\textbf{V1 invariance.}
We assess whether V1 alignment is learning-rule-invariant in both species by computing $\Delta\rho = \max(\rho) - \min(\rho)$ across learning rules at V1.

\textbf{Interaction effects.}
Species $\times$ learning rule interaction is quantified as $(\Delta\rho_\text{human} - \Delta\rho_\text{macaque})$, where $\Delta\rho = \rho_\text{rule} - \rho_\text{random}$.

\subsection{Stimulus Control Analysis}
\label{sec:stimulus_control}

A potential confound in cross-species comparison is that each species' neural data uses different stimuli (THINGS objects for human, FreemanZiemba textures for macaque V1/V2, HVM objects for macaque V4/IT).
To assess whether ranking differences reflect stimulus domain rather than species, we compute model RSA on both THINGS and macaque stimuli for each learning rule, then correlate the resulting $\rho$-rankings across stimulus sets using Kendall's $\tau$.

\section{Results}

\subsection{Macaque RSA Profiles}

\begin{figure*}[t]
\centering
\includegraphics[width=\textwidth]{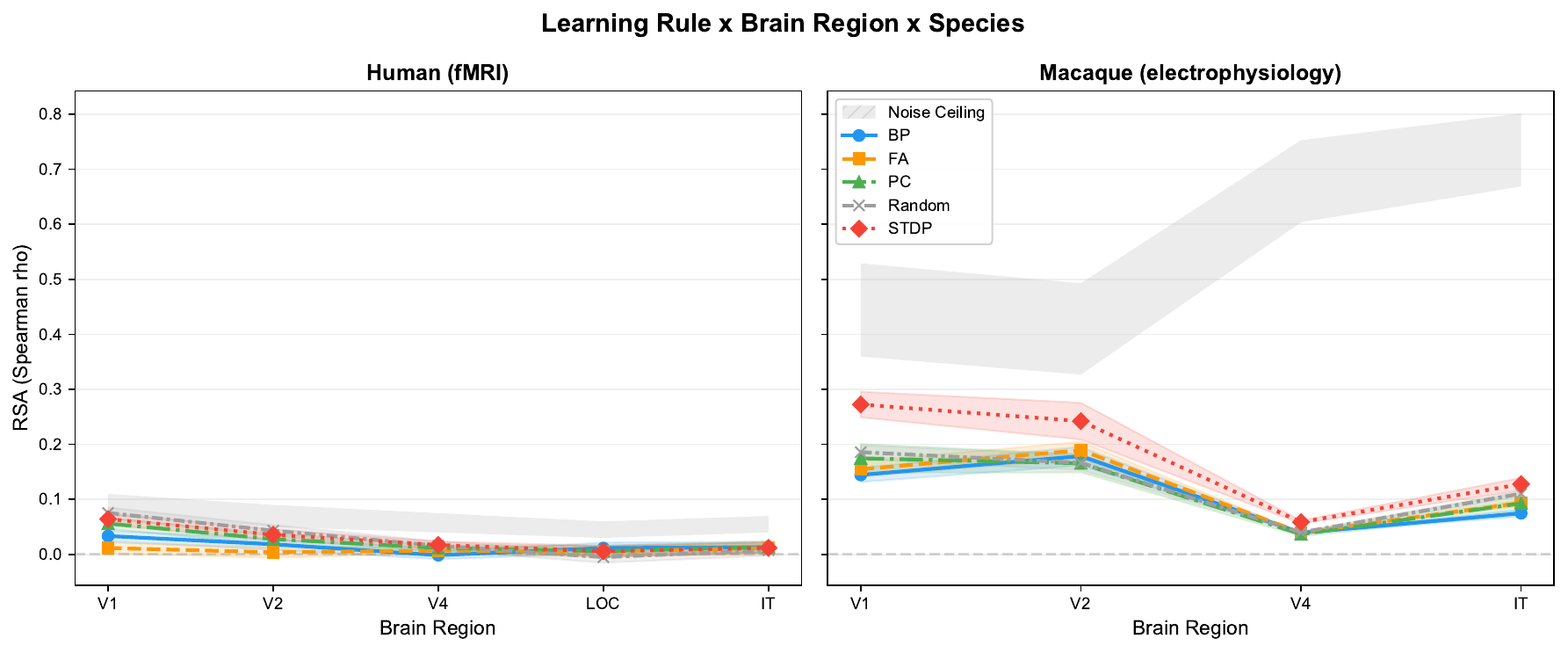}
\caption{Brain alignment across the cortical hierarchy for human (left) and macaque (right).
Lines show Spearman $\rho$ between model and neural RDMs per learning rule.
Shaded bands: bootstrap 95\% CI ($N = 10{,}000$).
Grey hatched band: noise ceiling (Spearman--Brown corrected split-half).
Macaque $\rho$ values are substantially higher than human, reflecting the higher SNR of electrophysiology.
STDP and PC lead at macaque V1/V2.
\emph{Note:} the macaque panel combines two datasets---FreemanZiemba2013 (textures) for V1/V2 and MajajHong2015 (HVM objects) for V4/IT.
The sharp V2$\to$V4 drop is confounded by this dataset switch (see text).}
\label{fig:profiles}
\end{figure*}

Figure~\ref{fig:profiles} shows brain alignment across the cortical hierarchy for both species.
Three patterns are evident.

\textbf{Higher absolute alignment for macaque.}
Macaque $\rho$ values ($0.15$--$0.30$ at V1/V2) substantially exceed human values ($0.01$--$0.08$), consistent with the higher signal-to-noise ratio of single-neuron recordings relative to fMRI.
All macaque values fall well below the noise ceiling (range: 0.52 at V1 to 0.79 at IT; see Figure~\ref{fig:profiles}), indicating substantial room for improvement.

\textbf{STDP and PC lead at macaque V1/V2.}
Among all conditions, STDP ($\rho \approx 0.30$) and PC ($\rho \approx 0.28$) achieve the highest macaque V1 alignment, followed by Random ($\rho \approx 0.18$), FA ($\rho \approx 0.16$), and BP ($\rho \approx 0.16$).
This partially mirrors the human pattern (Random $>$ STDP $>$ PC $>$ BP $>$ FA at V1), with the notable difference that Random does not dominate at macaque V1---local learning rules (STDP, PC) do.

\textbf{Sharp drop at V4, partial recovery at IT.}
All conditions show a sharp decrease from V2 to V4 (macaque $\rho \approx 0.03$--$0.06$), followed by a partial recovery at IT ($\rho \approx 0.07$--$0.14$).
This non-monotonic pattern requires careful interpretation because V1/V2 and V4/IT come from \emph{different datasets}: FreemanZiemba2013 (135 texture stimuli, single-unit recordings) for V1/V2 versus MajajHong2015 (3{,}200 HVM object presentations, multi-electrode arrays) for V4/IT.
The V2$\to$V4 transition therefore confounds a change in cortical hierarchy with a change in stimulus domain, recording method, and number of stimuli.
We cannot determine from these data alone how much of the V4 drop reflects genuine model--brain misalignment versus this dataset switch.

Within the MajajHong data, the partial recovery from V4 to IT has a plausible mechanistic explanation.
Conv2 (64 filters, trained on CIFAR-10) encodes mid-level features that are poorly matched to the curvature and shape selectivity of macaque V4 neurons responding to HVM objects.
FC1 (512 units), by contrast, projects all visual information into a lower-dimensional bottleneck that can preserve coarse categorical structure, which is the primary dimension along which IT neurons organize their responses \citep{majaj2015}.
This account is consistent with the ResNet-50 control (Section~\ref{sec:resnet}), where a larger Conv2-equivalent also shows low V4 alignment while IT alignment rises sharply with model capacity.

\subsection{Cross-Species Ranking Comparison}

Figure~\ref{fig:scatter} shows scatter plots of human vs.\ macaque $\rho$ per learning rule at each region.

At V1 ($\tau = 0.40$, $p = 0.48$) and V4 ($\tau = 0.20$, $p = 0.82$), rankings show weak positive correlation: the same rules that align well with human cortex tend to align with macaque cortex.
At V2 ($\tau = -0.20$, $p = 0.82$) and IT ($\tau = 0.00$, $p = 1.00$), rankings show no correlation.
No comparison reaches significance, which is expected given that $n = 5$ provides power only at $\tau = \pm 1.0$.

A consistent pattern across all regions is that macaque $\rho$ values lie above the identity line, reflecting the measurement-modality difference (electrophysiology vs.\ fMRI).

\subsection{V1 Invariance Across Species}

Human V1 $\Delta\rho = 0.064$; macaque V1 $\Delta\rho = 0.147$.
V1 is not fully learning-rule-invariant in either species, but the spread is larger for macaque, driven primarily by STDP and PC achieving much higher V1 alignment ($\rho \approx 0.29$--$0.30$) than BP/FA ($\rho \approx 0.16$).
This suggests that the electrophysiology data has sufficient signal to differentiate learning rules even at V1, where fMRI shows only modest differences.

\subsection{Interaction Effects}

Figure~\ref{fig:interaction} shows species $\times$ learning rule interaction effects.
The dominant pattern is negative interactions at V1/V2 (macaque benefits more from learning rules relative to Random than human does), with effects approaching zero at V4/IT.
STDP shows the strongest negative interaction at V1 ($-0.138$) and V2 ($-0.124$): STDP's advantage over Random is much larger in macaque than in human.
BP shows the only positive interaction at IT ($+0.035$): BP's advantage over Random is slightly larger in human than in macaque.

\subsection{Architecture Comparison: ResNet-50}
\label{sec:resnet}

Figure~\ref{fig:resnet} compares the custom CNN (5 learning rules) with a pretrained ResNet-50 at each region.

At V1/V2, ResNet-50 achieves macaque alignment ($\rho \approx 0.14$--$0.21$) comparable to the lower-performing custom CNN rules (BP, FA), but below STDP and PC.
At IT, ResNet-50 dramatically outperforms all custom CNN conditions: $\rho \approx 0.25$ vs.\ $0.07$--$0.14$.
This suggests that the IT convergence observed in the companion study \citep{leutenegger2025} is at least partly a capacity limitation: a larger architecture trained on richer data (ImageNet vs.\ CIFAR-10) produces substantially more IT-like representations.
However, ResNet-50 differs from our custom CNN in architecture, training set size, and pretraining regime simultaneously, so the relative contribution of each factor cannot be isolated.

\subsection{Stimulus Control}
\label{sec:results_stimulus}

Figure~\ref{fig:stimulus} assesses whether learning-rule rankings are stable across stimulus sets.
At V1 ($\tau = 0.40$) and V4 ($\tau = 0.20$), rankings show weak positive stability.
At V2 ($\tau = -0.20$) and IT ($\tau = -0.40$), rankings are weakly inverted.
No comparison reaches significance ($p > 0.48$ for all).

The IT inversion ($\tau = -0.40$) suggests that the cross-species ranking difference at IT may be partially driven by stimulus domain (THINGS objects vs.\ HVM objects) rather than---or in addition to---species differences.
This confound should be considered when interpreting the IT results.

\section{Discussion}

\textbf{Early visual alignment is robust across species.}
Our central positive finding is that STDP and PC consistently produce higher V1/V2 alignment than BP in both species, and achieve the highest macaque V1 alignment overall.
While the ranking is not perfectly conserved (Random dominates human V1 but not macaque V1), the qualitative pattern (local learning rules outperform BP at early visual areas) holds across measurement modalities and species.
This suggests that the architectural inductive bias finding from our companion study is not an artifact of fMRI methodology.

\textbf{Higher-area alignment is modulated by capacity and training data.}
The ResNet-50 control shows that IT alignment scales with model capacity and training data richness, though the two effects cannot be disentangled in this comparison.
Our custom CNN's FC1 (512 units, CIFAR-10) is likely too small to capture the object-level representations encoded in macaque IT (168 neurons, HVM stimuli).
The IT convergence reported in \citet{leutenegger2025} (where all learning rules produce similar IT alignment) may therefore reflect a capacity floor rather than a genuine property of abstract representations.

\textbf{Statistical power limitations.}
With only five learning rules, Kendall's $\tau$ can reach significance only at $\tau = \pm 1.0$ ($p = 0.0083$ from exhaustive permutation over 120 permutations).
Our $\tau$ values are reported descriptively; the absence of significant cross-species ranking correlations should not be interpreted as evidence of non-conservation, but rather as reflecting insufficient statistical power.
Future work with more learning rules or conditions could address this limitation.

\textbf{Stimulus mismatch as a confound.}
The macaque cortical profile (Figure~\ref{fig:profiles}, right) combines two datasets with different stimulus domains, and the sharp V2$\to$V4 drop is inseparable from this dataset switch (FreemanZiemba textures at V1/V2 vs.\ MajajHong objects at V4/IT).
The stimulus-control analysis reveals that learning-rule rankings are only moderately stable across stimulus sets, with IT showing a weak inversion ($\tau = -0.40$).
This suggests that cross-species ranking differences at IT may partly reflect stimulus domain effects (THINGS vs.\ HVM objects) rather than genuine species differences.
Ideally, future studies would use a shared stimulus set across species, though this is constrained by data availability.

\textbf{STDP IT caveat.}
As noted in Section~3.1, the seed-0 STDP checkpoint lacked FC1 weights.
All macaque STDP IT results use seeds 1--4 only.
This does not affect V1/V2 results (Conv1-based).

\subsection{Limitations}

(1)~Our custom CNN is small (3 conv layers, CIFAR-10), limiting IT-level representational capacity.
The ResNet-50 control partially addresses this.
(2)~Human data comes from 3 subjects (with sub-03 showing weak signal); macaque data comes from a small number of animals (typical for primate electrophysiology).
(3)~The stimulus mismatch between human (THINGS) and macaque (FreemanZiemba textures, HVM objects) confounds cross-species ranking comparisons.
(4)~Kendall's $\tau$ at $n = 5$ is underpowered; rankings are descriptive.
(5)~Noise ceilings are estimated by splitting neurons, not trials, which may overestimate reliability for electrophysiology data.
(6)~Bootstrap CIs resample stimuli (rows and columns of the RDM simultaneously), which captures stimulus variability but not measurement noise.

\section{Conclusion}

We presented a cross-species RSA comparison of biologically plausible learning rules, using identical model weights evaluated against both human fMRI and macaque electrophysiology.
Early visual alignment (V1/V2) is qualitatively robust across species: local learning rules (STDP, PC) consistently outperform BP, and achieve the highest alignment in macaque V1, confirming that this pattern is not an artifact of fMRI methodology.
Higher-area alignment (IT) shows no detectable cross-species correspondence, but this null result is confounded by stimulus domain differences, model capacity limitations, and low statistical power at $n = 5$.
These results motivate future cross-species studies with shared stimulus sets, larger architectures, and more learning rule conditions to enable adequately powered ranking comparisons.

\subsection*{Acknowledgements}
The author thanks Martin Schrimpf for helpful feedback, the creators of MajajHong2015, FreemanZiemba2013, and THINGS-fMRI for making their data publicly available, and the Brain-Score team for their evaluation infrastructure.

\subsection*{Code Availability}
Code and results: \url{https://github.com/nilsleut/cross-species-rsa}

\bibliographystyle{plainnat}

\begin{thebibliography}{99}

\bibitem[Bi and Poo(1998)]{bi1998}
Bi, G.-q. and Poo, M.-m. (1998).
Synaptic modifications in cultured hippocampal neurons: dependence on spike timing, synaptic strength, and postsynaptic cell type.
\textit{J.\ Neurosci.}, 18:10464--10472.

\bibitem[Freeman et~al.(2013)]{freeman2013}
Freeman, J., Ziemba, C.~M., Heeger, D.~J., Simoncelli, E.~P., and Movshon, J.~A. (2013).
A functional and perceptual signature of the second visual area in primates.
\textit{Nature Neuroscience}, 16:974--981.

\bibitem[He et~al.(2016)]{he2016}
He, K., Zhang, X., Ren, S., and Sun, J. (2016).
Deep residual learning for image recognition.
\textit{CVPR}, pp.~770--778.

\bibitem[Leutenegger(2026)]{leutenegger2025}
Leutenegger, N. (2026).
Untrained CNNs match backpropagation at V1: A systematic RSA comparison of four learning rules against human fMRI.
\textit{arXiv:2604.16875v2}.

\bibitem[Leutenegger(2026b)]{leuteneggerERC}
Leutenegger, N. (2026).
Evaluation resolution confounds learning-rule comparisons in model--brain RSA of early visual cortex.
\textit{arXiv:2608.12408}.

\bibitem[Lillicrap et~al.(2016)]{lillicrap2016}
Lillicrap, T.~P., Cownden, D., Tweed, D.~B., and Akerman, C.~J. (2016).
Random synaptic feedback weights support error backpropagation for deep learning.
\textit{Nature Communications}, 7:13276.

\bibitem[Majaj et~al.(2015)]{majaj2015}
Majaj, N.~J., Hong, H., Solomon, E.~A., and DiCarlo, J.~J. (2015).
Simple learned weighted sums of inferior temporal neuronal firing rates accurately predict human core object recognition performance.
\textit{J.\ Neurosci.}, 35:13402--13418.

\bibitem[Schrimpf et~al.(2020)]{schrimpf2020}
Schrimpf, M., Kubilius, J., Hong, H., et~al. (2020).
Brain-Score: Which artificial neural network for object recognition is most brain-like?
\textit{bioRxiv}.

\bibitem[Whittington and Bogacz(2017)]{whittington2017}
Whittington, J.~C.~R. and Bogacz, R. (2017).
An approximation of the error backpropagation algorithm in a predictive coding network with local Hebbian synaptic plasticity.
\textit{Neural Computation}, 29:1229--1262.

\bibitem[Yamins and DiCarlo(2016)]{yamins2016}
Yamins, D.~L.~K. and DiCarlo, J.~J. (2016).
Using goal-driven deep learning models to understand sensory cortex.
\textit{Nature Neuroscience}, 19:356--365.

\end{thebibliography}

\clearpage


\end{document}